\documentclass[conference]{IEEEtran}
\IEEEoverridecommandlockouts
\usepackage{cite}
\usepackage{amsmath,amssymb,amsfonts}
\usepackage{url}
\usepackage{algorithmic}
\usepackage{graphicx}
\usepackage{tabularx}
\usepackage{pgfplots}
\usepackage{textcomp}
\usepackage{xcolor}
\usepackage{multirow}
\def\BibTeX{{\rm B\kern-.05em{\sc i\kern-.025em b}\kern-.08em
    T\kern-.1667em\lower.7ex\hbox{E}\kern-.125emX}}

\makeatletter % changes the catcode of @ to 11
\newcommand{\linebreakand}{%
  \end{@IEEEauthorhalign}
  \hfill\mbox{}\par
  \mbox{}\begin{@IEEEauthorhalign}
}
\makeatother % changes the catcode of @ back to 12

\usepackage[font=footnotesize]{subcaption}
\usepackage[font=footnotesize]{caption}
\usepackage{balance}

\let\oldtextcolor\textcolor
\renewcommand{\textcolor}[2]{\oldtextcolor{black}{#2}}

\let\oldcolor\color
\renewcommand{\color}[1]{\oldcolor{black}}

\begin{document}

% \title{\LARGE \bf Optimising Remote Error Resolution with Virtual and Mixed Reality}

\title{Mixed Reality Outperforms Virtual Reality for Remote Error Resolution in Pick-and-Place Tasks}

% \author{Advay Kumar$^{1,*}$, Stephanie Simangunsong$^{1,*}$, Pamela Carreno-Medrano$^{1}$, Akansel Cosgun$^{2, \dagger}$%
% \thanks{$^{1}$Monash University, Australia}
% \thanks{$^{2}$Deakin University, Australia}
% \thanks{$^{*}$Equal Contribution}
% \thanks{$^{\dagger}$Corresponding author {\tt\footnotesize akan.cosgun@deakin.edu.au}}}

\author{\IEEEauthorblockN{Advay Kumar$^{*}$}
\IEEEauthorblockA{
\textit{Faculty of Engineering} \\
\textit{Monash University}\\
Clayton, Australia \\
adva0001@student.monash.edu}
\and
\IEEEauthorblockN{Stephanie Simangunsong$^{*}$}
\IEEEauthorblockA{\textit{Faculty of Engineering} \\
\textit{Monash University}\\
Clayton, Australia \\
ssim0033@student.monash.edu}

\and
\IEEEauthorblockN{Pamela Carreno-Medrano}
\IEEEauthorblockA{\textit{Faculty of Engineering} \\
\textit{Monash University}\\
Clayton, Australia \\
pamela.carreno@monash.edu}

\linebreakand
\and
\IEEEauthorblockN{Akansel Cosgun$^{\dagger}$%}
\IEEEauthorblockA{\textit{Faculty of Science Engineering and Built Environment} \\
\textit{Deakin University}\\
Melbourne, Australia \\
akan.cosgun@deakin.edu.au}}

\thanks{$^{*}$Equal Contribution}
\thanks{$^{\dagger}$Corresponding author }
% reduce space between author block and content
\\[-6.5ex]
}

\maketitle
% Adjust space between figure caption
\setlength{\belowcaptionskip}{-4pt}
% Adjust space after figure labelling
\setlength{\textfloatsep}{5pt}
% Adjust space between section heading
\setlength{\floatsep}{-5pt}

%original submission
%\begin{abstract}
%The deployment of robotic systems in manufacturing and warehouse settings is rapidly increasing, particularly with articulated robots being used for packaging tasks. Despite advancements in AI and robotics, these systems often require human intervention to resolve errors, raising safety concerns and causing downtime for adjacent robots. One promising solution is remote operation for error resolution. However, traditional methods, such as 2D camera streams, have limitations. Operators frequently struggle to communicate high-level goals to robots, and robots often fail to provide adequate feedback on their operational status. This paper explores the potential of virtual reality (VR) and mixed reality (MR) technologies for improving remote error resolution and addressing the shortcomings of traditional methods. We present a user study comparing VR, MR, and conventional 2D camera streams in resolving common robot errors in distribution centres. Our results indicate that MR, which presents a combination of the physical and virtual world, was preferred over VR and Camera Streams in terms of perceived interface usability and comfort. By quantitatively and qualitatively comparing immersive technologies and traditional methods for remote operation, this work advances the field by showcasing how immersive technologies can enhance human-robot interaction and improve the safety and efficiency of automated systems in warehouses and distribution centres.
%\end{abstract}

% AC version
\begin{abstract}
This study evaluates the performance and usability of Mixed Reality (MR), Virtual Reality (VR), and camera stream interfaces for remote error resolution tasks, such as correcting warehouse packaging errors. Specifically, we consider a scenario where a robotic arm halts after detecting an error, requiring a remote operator to intervene and resolve it via pick-and-place actions. Twenty-one participants performed simulated pick-and-place tasks using each interface. A linear mixed model (LMM) analysis of task resolution time, usability scores (SUS), and mental workload scores (NASA-TLX) showed that the MR interface outperformed both VR and camera interfaces. MR enabled significantly faster task completion, was rated higher in usability, and was perceived to be less cognitively demanding. Notably, the MR interface, which projected a virtual robot onto a physical table, provided superior spatial understanding and physical reference cues. Post-study surveys further confirmed participants’ preference for MR over other interfaces.
\end{abstract}

% \begin{IEEEkeywords}
% Teleoperation, human-robot interaction, robot mistakes, robot error correction, virtual reality, mixed reality, interface design
% \end{IEEEkeywords}

\section{Introduction and Related Works}
%Robotic systems have been increasingly incorporated into a plethora of industries such as healthcare, mining, defence and logistics \cite{9911168} \cite{7237701} \cite{drones7020080}. 

Over the past decade, the rise of online shopping has driven the integration of robotic systems into logistics operations, a trend further accelerated by the COVID-19 pandemic and labour shortages \cite{shaw2022online}. Mobile robots are commonly used to transport goods within warehouses, while robotic manipulators, which is our focus, are vital for packaging and sorting tasks in distribution centers \cite{amazonWebsite}. Notable companies like Amazon \cite{amazon-sparrow}, Ocado \cite{ocado}, and Walmart \cite{walmart} have adopted this technology. Despite advancements in robotic manipulation, including the use of AI for decision-making \cite{8895161}, these systems still encounter frequent operational challenges \cite{honig2018understanding}, such as technical issues and interaction failures. Addressing these errors often requires direct human intervention, raising safety concerns due to the need to stop the robot or place workers in potentially hazardous environments \cite{8832130}. %In these facilities, processes such as picking, packaging, sorting, storing, and labelling goods are becoming increasingly automated with the help of robot manipulators, a type of multi-jointed pick-and-place robot that mimics the motion of a human arm. 
%Notable examples of companies deploying this technology include Amazon \cite{amazon-sparrow}, Ocado \cite{ocado}, and Walmart \cite{walmart}. 

%While robot manipulators are becoming increasingly capable, with some even incorporating advanced decision-making capabilities through Artificial Intelligence (AI)~\cite{8895161}—such as the Sparrow robot used by Amazon \cite{amazon-sparrow}—these robotic arms still encounter operational challenges \cite{honig2018understanding} including frequent failure caused by both technical issues and interaction failures. Addressing these challenges often necessitates direct physical human assistance, which introduces numerous safety and efficacy concerns, such as the need to stop the robot and surrounding machinery or having a person enter a potentially hazardous environment with industrial equipment \cite{8832130}.

Remote control of robot arms provides a safer alternative, allowing operators to resolve errors without being physically present. Previous works have demonstrated the potential of remote control in addressing situational errors \cite{wozniak2023happily}, which are errors caused by the robot's environment or circumstances, such as object placement, blocked objects, or the robot's camera being partially obstructed. While traditional teleoperation methods, like desktop interfaces, allow information exchange between humans and robots, they often lack the contextual awareness needed to fully understand and resolve issues within the robot's environment \cite{dragan2013policy}. Achieving intuitive and effective remote control remains an open challenge. 

Recent advancements in consumer-grade Virtual, Augmented and Mixed Reality (VR, AR and MR) headsets led to the proliferation of applications for Human-Robot Interaction (HRI), mainly for teleoperation \cite{xr-review,10.1145/3597623}. Research has shown that VR interfaces perform better than desktop interfaces in terms of both task performance and user experience \cite{whitney2019comparing, gui-vs-vr, shi2023research}. Nakamura et. al. \cite{nakamura2020dual} developed a dual-arm control interface where one monitor showed camera views and another monitor showed a virtual simulation. They found that the virtual display reduced task time and collisions. Waymouth et. al. \cite{waymouth2021demonstrating} similarly showed two monitors for a cloth manipulation task, and compared this desktop interface with AR. They found that AR was found more suitable for understanding the task while the desktop interface was better for repetition. Ulloa et al. \cite{sar-teleop} compared MR with a desktop interface for a quadruped with a robotic arm, and found that MR resulted in increased user confidence and improved task metrics. \textcolor{blue}{In a separate study, Ulloa et al. \cite{cruz2024analysis} analyzed VR teleoperation in a remote setting and MR teleoperation in an in-situ setting for the same type of robot. They found VR to be more effective overall, offering users greater security compared to MR. However, MR supported decision-making and reduced user stress levels by allowing users to remain in the same physical area as the robot.}

Extended reality (XR) technologies, which include VR and MR, have also been applied to diagnosing robot errors. Avalle et al. \cite{adaptive-ar} used AR to identify faults, showing that adaptive mobility, where error hints adjust to user movements, led to faster fault identification. Kaipa et al. \cite{robot-preception-error} explored remote human-robot collaboration for resolving perception errors, demonstrating improvements in task performance. Wozniak et al. \cite{wozniak2023happily} showed that users preferred VR over traditional methods for correcting perception errors.

%AR/VR/MR have also been employed to assist users in diagnosing and detecting robotic faults \cite{adaptive-ar,robot-preception-error, wozniak2023happily}. Avalle et al. \cite{adaptive-ar} used AR to help identify faults and compared adaptive mobility, where error hints adjust to user movement and robot placement, with non-adaptive approaches. Adaptive mobility led to quicker fault identification. Similarly, Kaipa et al. \cite{robot-preception-error} explored human-robot collaboration for correcting perception errors, finding that remote operator assistance improved task performance. Wozniak et al. \cite{wozniak2023happily} developed a VR-based framework for correcting robotic perception errors, showing that users preferred VR and improved more quickly with it compared to traditional interfaces.

However, gaps in the literature remain. First, few studies focus on how AR/VR/MR technologies can assist with robotic fault resolution rather than just diagnosis \cite{adaptive-ar}. \textcolor{blue}{Second, although most research compares XR interfaces to desktop setups, little attention has been given to whether VR, AR, or MR is better suited for performing identical teleoperation tasks under the same conditions, whether in-situ or remote.} Third, comparisons often conflate differences in visualization elements with differences in form factor (e.g., desktop vs. headsets). To address these gaps, this paper presents a user study that compares VR, MR, and camera-only interfaces for remote error resolution in robotic manipulation tasks. Our contributions are as follows:

\begin{enumerate}
\color{blue}
    \item An empirical comparison of VR and MR against camera-only interfaces for telemanipulation in the context of error resolution, where participants are informed of the error type and use the interfaces to resolve it.
    \item A novel comparison using the same headset and controller across VR, MR, and camera streams for the same telemanipulation task and in the same settings.
    \item A detailed discussion on the implications of using the same device, control setup, and environment when comparing multiple interfaces.
\end{enumerate}

\section{Methodology}
This section will outline the error scenario utilized in the study, detailing its implementation and application within the experimental setup. 

\subsection{Error Scenario}
In warehouse environments, articulated robots commonly perform tasks like sorting and packaging, where they pick objects and place them in designated bins. During these processes, various errors can occur. Mitash et al. \cite{10160846} introduced a dataset for robotic pick-and-place activities, including two key types of robot-induced errors: (1) multi-pick errors, where multiple objects are picked by mistake, and (2) package errors, where an object's packaging is damaged, causing the object to break into parts. Our chosen error scenario simulates the latter, where a robot inadvertently scatters and drops objects around the workstation. Resolving this error involves scanning the area to locate all objects, assessing their condition, and determining the appropriate action—either discarding damaged items or placing intact objects in a packaging bin. While the robot can detect the error, we assume it either lacks the ability to fix it autonomously or would take too long to do so, making human intervention necessary. A remote operator connects to the robot via a head-mounted display, tasked with controlling the robotic arm to execute pick-and-place actions and resolve the error.

\subsection{Scenario Implementation}

\begin{figure}[ht!]
    \centering
    \includegraphics[width=1\linewidth]{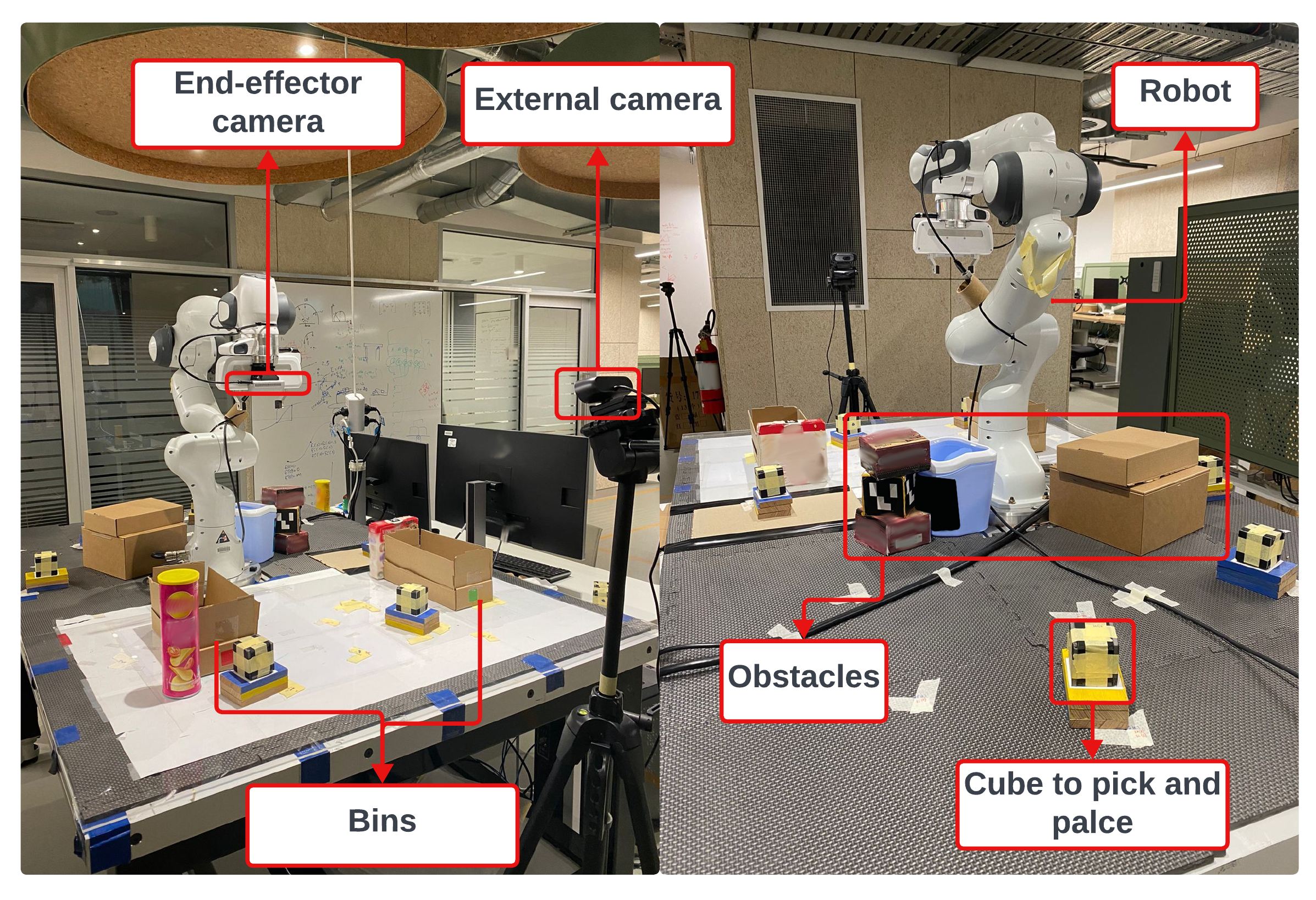}
    \caption{Scenario used in the study including the robot, cameras, bins and cubes}
    \label{fig:scenario}
\end{figure}

A depiction of this scenario as presented to participants (also referred to as operators) in our user study is shown in Fig. \ref{fig:scenario}. Six cubes were scattered around the robot’s surroundings, each marked on the bottom with either a red or green sticker. Two bins were present, one is a packaging bin (labelled green) and the other is a discard bin (labelled red). The labels represented ``good" (green) and ``damaged" (red). Stickers were used to simplify the task. To make the scenario more interesting, additional objects were placed to serve as obstacles and partially block some of the scattered cubes. 

Operators were provided with visibility of the robot’s environment through two cameras. One was the end-effector (EE) camera, attached to the robot's end-effector, which had object detection capability. The other was an external camera, without object detection capability. The robot was only aware of the EE camera and could only detect what this camera saw. The inclusion of the EE camera aligns with findings from previous studies \cite{vr-camera}, which have demonstrated that mounting a camera on the robot's end-effector improves the operator's visualization of the remote scene, leading to enhanced task performance. Meanwhile, the external camera provided the human operator with additional context about the robot's surroundings, facilitating the identification and localization of objects, such as cubes, within the workspace.

The operator could manipulate the robot's first and sixth joints to adjust the EE camera’s position via a joystick. The first joint, located at the base of the robot, enabled it to rotate clockwise and counterclockwise around the workstation. The sixth joint allowed the end-effector to tilt forward and backward (note that the Panda robot is a 7 DoF arm). Together, these joints enabled users to explore the robot's workstation and gain better visibility of the environment. A click-and-go operation was implemented to execute the pick-and-place task using the same controller. This operation allowed the operator to click on a cube, instructing the robot to pick it up. Once the robot successfully grasped the cube, the operator could click on the desired bin to place the cube. We note that the robot remained stationary until it received instructions from the human operator. Despite previous studies suggesting that waypoint control for teleoperation performs better for novice users \cite{9106630}, the decision to use the click-and-go mechanism in this study was driven by its simplicity. The error correction tasks presented here did not require a complex teleoperation procedure but instead involved straightforward pick-and-place actions. By using the click-and-go method, participants could focus solely on selecting which objects to pick and place, without needing to manage the robot’s grasping position.

Given this scenario implementation, a participant would need to complete the following steps to solve the robot error:
\begin{enumerate}
    \item Explore the robot's environment and locate a scattered cube using the EE camera. This involves moving the robot's 1st joint horizontally.
    \item Pick up the scattered cube using the click-and-go operation.
    \item Inspect the cube held by the robot to determine the cube's condition. This inspection process requires moving the robot's sixth joint forward to reveal the sticker on the bottom of the cube in the external camera. 
    \item Place the cube in the appropriate bin using the click-an-go operation.
    \item Repeat this process until all cubes are found.
\end{enumerate}

\subsection{Visualisation}
We developed three different interfaces for this study: VR, MR, and camera streams. All interfaces employed the EE and external cameras and used a slightly different click-and-go mechanism. For VR and MR, users clicked on the virtual cubes while for camera streams users clicked directly on the EE camera video stream. Similarly, the same control approach for the robot's first and sixth joints was used across all interfaces. 

\begin{figure}
     \centering
     \begin{subfigure}[b]{0.85\linewidth}
         \centering
         \includegraphics[width=\textwidth]{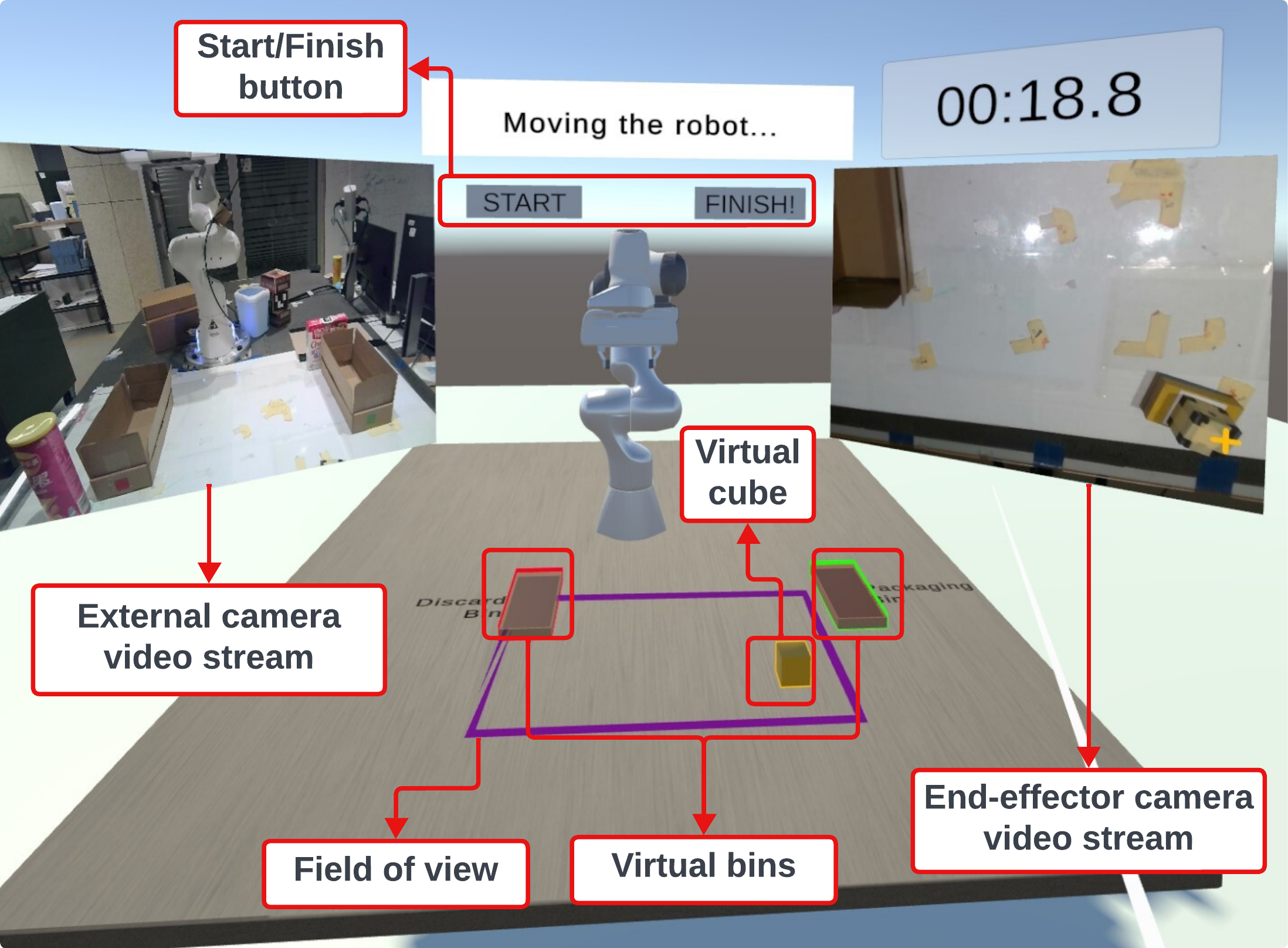}
         \caption{\textcolor{blue}{Virtual Reality (VR) Interface}}
         \vspace{3pt}
         \label{fig:VR}
     \end{subfigure}
     \hfill
     \begin{subfigure}[b]{0.85\linewidth}
         \centering
         \includegraphics[width=\textwidth]{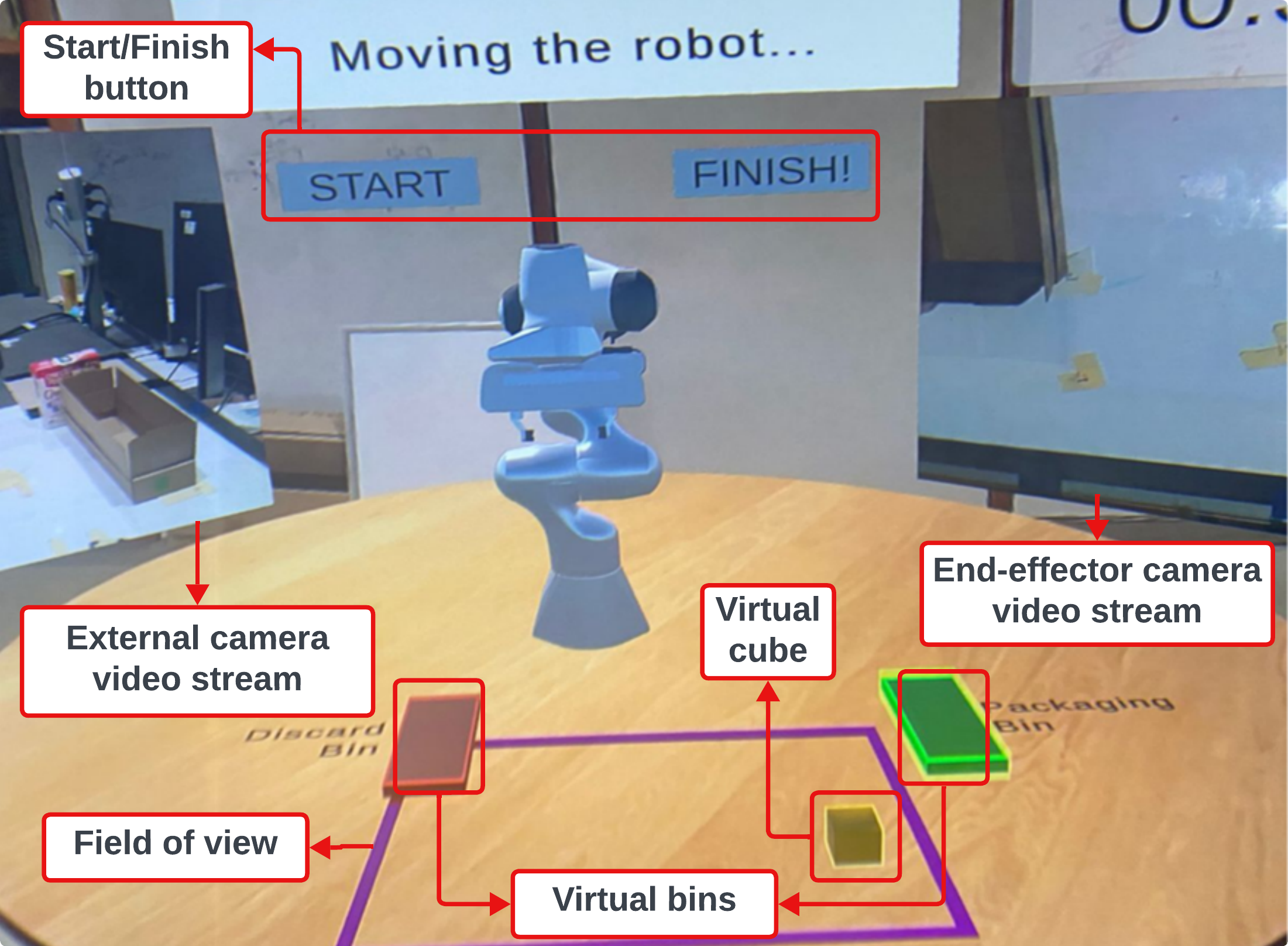}
         \caption{\textcolor{blue}{Mixed Reality (MR) Interface}}
         \vspace{3pt}
         \label{fig:MR}
     \end{subfigure}
     \hfill
     \begin{subfigure}[b]{0.85\linewidth}
         \centering
         \includegraphics[width=\textwidth]{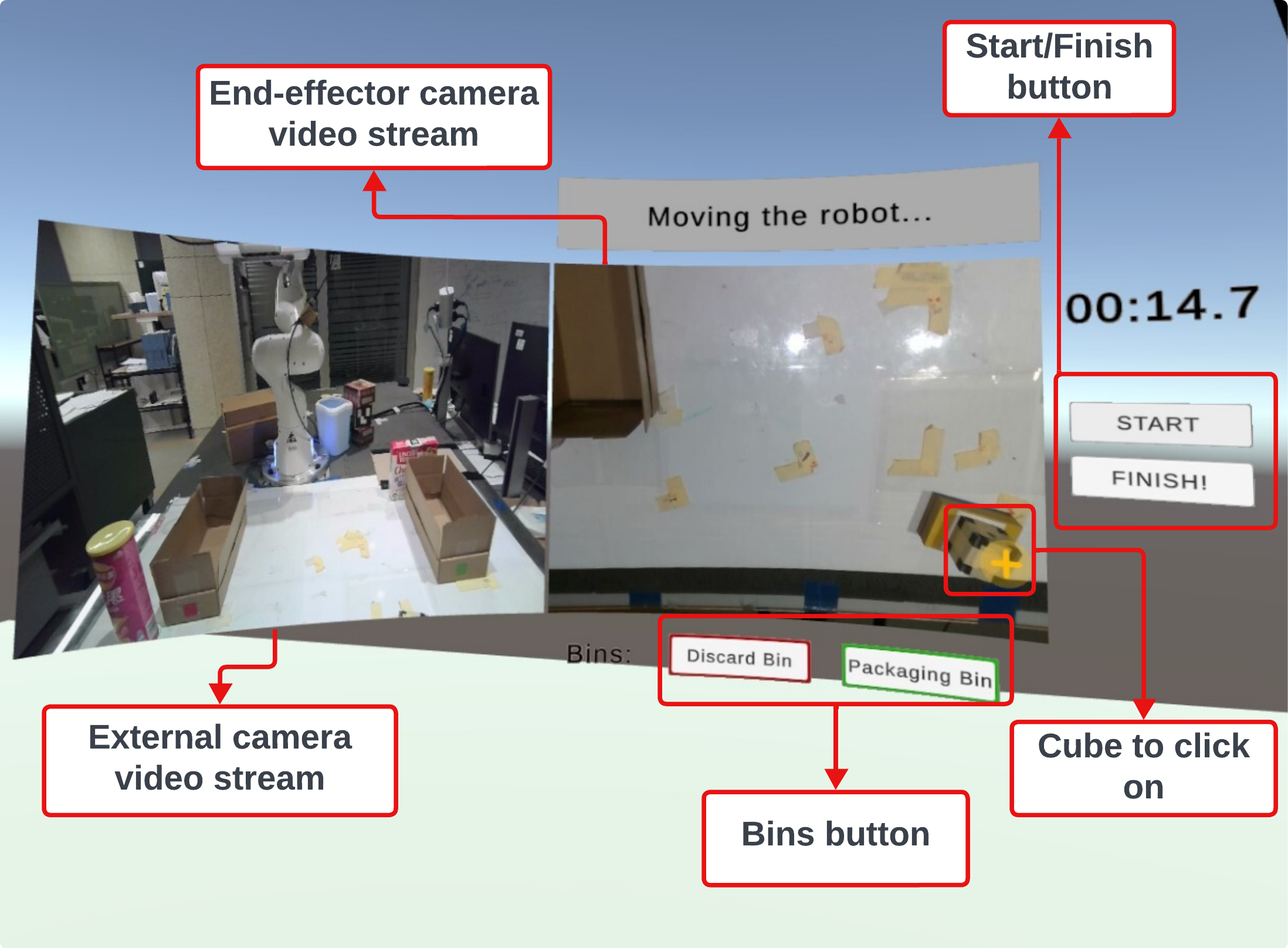}
         \caption{Camera Streams Interface}
         \vspace{3pt}
         \label{fig:camera stream}
     \end{subfigure}
        \caption{Visualisations presented to participants during the user study.}
        \label{fig:visualisation}
\end{figure}

\subsubsection{Virtual Reality (VR)}
In this interface \textcolor{blue}{(see Fig. \ref{fig:VR})}, users were fully immersed in a virtual environment. This environment consisted of a virtual copy of the robot's environment, including a virtual table, robot, bins and detected cubes. The robot's field of view—indicating which parts of the table were visible to the EE camera—was drawn on the table, providing users with additional context. To interact, users needed to click on the virtual objects to pick them up and then click on the virtual bins to place the objects.

\subsubsection{Mixed Reality (MR)}
In this interface \textcolor{blue}{(see Fig. \ref{fig:MR})}, users could perceive their surroundings through the pass-through feature, allowing them to view the physical world. This interface presented similar information to that of VR, with the primary difference being that MR used the user's physical table, whereas VR utilized a virtual table.

\subsubsection{Camera Streams}
In this interface, users were only able to see the video feeds from the two cameras (EE and external). There was no virtual copy of the robot's environment. Objects detected by the EE camera were marked with a simple cross marker and a translucent circle, as shown in Fig. \ref{fig:camera stream}. To pick up a cube, users had to click on the markings in the EE camera's video stream. Instead of virtual bins, two buttons were provided: one labelled "discard" with a red border and the other "packaging" with a green border. Users needed to click on these buttons to place the picked objects into the corresponding bins.

\section{Error Correction User Study}

We conducted a user study to evaluate whether participants' performance when resolving the robot error varied across the three different interfaces. In this section, we first outline the hypotheses of our study. We then describe the study setup and procedure and the metrics we collected for each participant. Finally, we report on the general characteristics of our participants. 

\subsection{Study Hypotheses}
We consider the following hypotheses for this study:

\noindent\textbf{H1:} \textit{Users will complete the task faster with the MR interface compared to the VR and camera interfaces}

The enhanced scene representation in MR and VR is likely to aid users in understanding which parts of the table are visible to the robot and the robot’s current pose. This may help them explore the robot's environment more effectively. Additionally, users are expected to feel more comfortable moving around in MR compared to VR, making them less cautious and enabling quicker error resolution. MR also allows users to view their own arms which research suggests helps with spatial understanding \cite{coello2005spatial}, this may help users complete the task faster.

\noindent\textbf{H2:} \textit{The MR interface will be rated  higher in usability and lower in perceived workload compared to the VR and camera interfaces}

We suspect the MR interface to have higher usability as it combines the real world and virtual environment, allowing for easier contextualisation and spatial reasoning. Users may be able to use the real table in front of them as an anchor point and more easily complete the task compared to the VR and camera interfaces where everything is virtual.

\noindent\textbf{H3:} \textit{The MR interface will be the most preferred interface}

Knowing the real world around them, users may feel more comfortable using the MR interface and prefer it over the VR and camera interfaces.

\subsection{Study Setup and Procedure}

The experiment was conducted in the Monash Robotics Centre, which was approved by the Monash University Human Research Ethics Committee (MUHREC), Project ID: 43579. \textcolor{blue}{The hardware and software was set up as described in Appendix A.} 
Participants stood in front of a table located in a different area from the robot to simulate a remote setting. 

To avoid repetition bias, three variations of the same error scenario were employed, differing in the placement of the cubes. This was introduced to ensure participants engaged with different contexts, enabling a more comprehensive assessment of how each visualization affected error resolution. To ensure equal difficulty across these variations, the robot workstation was divided into specific areas, and the number of objects placed in each area remained consistent across all variations. A test run done by a researcher, using the same interface for all variations was conducted to ensure that the time required to resolve each scenario was similar. The order in which participants encountered both the visualizations and error scenarios was randomized for each participant.

The study took approximately 65 minutes per participant, with a researcher present to observe and supervise. Before the study commenced, participants were asked to read the explanatory statement, sign a consent form and complete a demographic survey, collecting information such as age, gender, occupation, and prior experience with XR devices and robots, particularly in the context of HRI. The study consisted of two phases: training, designed to help participants learn and become familiar with the interfaces and testing, used to assess the interfaces' performance.

In the training phase, participants watched a training and safety induction video that introduced the scenario to participants and relayed information on how to use the control system. Participants were then given time to practice resolving a simplified error scenario with each interface to familiarize themselves with the system. During this time, participants could ask questions about the control system. The training phase concluded with participants watching a second video, where they were informed that the error-resolving operation was costly and that they would be timed to encourage timely resolution. 

In the testing phase, participants resolved the presented error using one of the interfaces. They could start their attempt by clicking the start button and conclude it by clicking the finish button, the button can be seen in each interface shown in Fig.~\ref{fig:visualisation}. 
Once an attempt concluded, the state of the scene at that point determined whether the attempt was successful or not. Information regarding how the success or failure of an attempt is \textcolor{blue}{determined} is provided in the Objective Metrics Section (Section \ref{objective metrics}). Afterwards, they were asked to fill out a post-interface questionnaire regarding their experience with that interface. This process was repeated for the remaining interfaces. 
Once the last post-interface questionnaire was finished, participants were asked to complete a post-study survey (Table \ref{post-study}). This survey provided participants with the opportunity to reflect on their experience and express their preferences regarding the error resolution interface/mode.  

\begin{table}
\centering
\caption{Post-Study Survey. These questions will be answered using a 5-point Likert scale unless otherwise specified.}
    \begin{tabular}{ |m{1em} m{27em}| }
    \hline
        1: & What is your preference among the three modes/interfaces for correcting robot errors?. \{Multiple-choice: VR, MR, Camera Streams.\} \\
        \hline
        2: & Based on your experience, which interface do you believe would be the most comfortable to use for an extended period of time? \{Multiple-choice: VR, MR, Camera Streams.\}\\
        \hline
        3: & (For all interfaces) The video stream from the end-effector camera was useful in understanding the robot's environment.\\
        \hline
        4: & (For all interfaces) The video stream from the external camera was useful in understanding the robot's environment.\\
        \hline
        5: & The click-and-go mechanism to pick and place objects was easy to use.\\
        \hline
        6: & The headset was comfortable to use.\\
        \hline
        7: & (Optional open-ended) How would you describe your experience?\\
        \hline
        8: & (Optional open-ended) Do you have any suggestions or comments regarding the experience?\\
        \hline
        9: & (Optional open-ended) Did you notice any type of discomfort while using the headset?\\
        \hline
        10: & (Optional open-ended) What is something that you would change in each interface?\\
    \hline
    \end{tabular}
    \label{post-study}
\end{table}

\subsection{Metrics}
\subsubsection{Objective Metrics}
\label{objective metrics}
To evaluate the performance of error resolution, specific criteria were established to determine a successful attempt by each participant. An attempt was considered successful if the participant rectified the robot error, meaning that all cubes had to be picked up and placed in the right bin based on their label: red-labelled cubes in the discard bin and green-labelled cubes in the packaging bin. If these were not met, the attempt was deemed unsuccessful. The final state of the scene at the end of each attempt was assessed based on these criteria.

To analyze the performance of the different interfaces, the following metrics were collected at the end of each attempt:
\begin{itemize}
    \item \textbf{Success of the attempt:} Whether the attempt was deemed successful or not based on the established criteria.
    \item \textbf{Resolution time:} The time from the start of the attempt (once the start button is clicked) to the end of the attempt (when the finish button is clicked).
    \item \textbf{Contextualization time:} The time taken for a participant to pick and place the first object. The measurement begins once the attempt starts (a participant clicks on the start button) and ends once the cube is placed in a bin.
    \item \textbf{Number of objects left:} The number of objects that were not picked up in an attempt.
\end{itemize}

If an attempt failed due to a technical issue (e.g., the robot failed to grasp the object or the object slipped from the gripper), the attempt was discarded. In such cases, the scenario was reset to its initial state, and the user was asked to try again. These metrics were used to evaluate the objective aspects of the study, specifically H1.

\subsubsection{Subjective Metrics}

After completing each scenario, participants were invited to fill out a post-interface questionnaire. This questionnaire consisted of two parts:
\begin{enumerate}
    \item \textbf{System Usability Scale (SUS)} \cite{sus}: was used to assess the usability of each interface. It was chosen for its simplicity and ability to provide a quick assessment of the user experience across various systems. This questionnaire was used to determine whether participants perceived one interface as easier to use than the others. The score obtained from the System Usability Scale (SUS) ranges from 0 to 100, with higher scores indicating higher perceived usability.
    \item \textbf{NASA Task Load Index (NASA TLX)} \cite{nasa}: was used to evaluate the task's feasibility and workload. It was chosen for its ability to effectively assess the workload of a given task, enabling comparison of task load across different interfaces. This allowed us to understand whether certain interfaces were more challenging or required more effort to use compared to others. The score obtained from the NASA Task Load Index (NASA TLX) ranges from 0 to 100, with lower scores indicating that the task required less effort, making the system easier to use in terms of workload.
\end{enumerate}

\subsection{Participants}
The study included a total of N = 21 participants, recruited through advertising across the university campus, faculty mailing lists, and hallway recruitment. All participants were university students or staff, aged between 19 and 28 (mean = 21.6 years, SD = 2.1 years). Of the 21 participants, two identified as female and 19 identified as male.

Regarding the experience with VR, AR, and MR devices, nine participants had no prior experience, 11 had minimal experience (seven hours or less), and one had extensive experience (30 hours or more). Participants were also asked about their experience interacting with robots. Seven reported having no prior experience, seven had minimal experience (seven hours or less), three had moderate experience, and four had extensive experience (30 hours or more).

\section{Results}

We used linear mixed models (LMMs) for our data analysis to account for the repeated measures design, where participants interacted with all interfaces across different scenarios. 
We used R \cite{RProject} with the lme4 package by Bates, Maechler and Bolkar \cite{JSSv067i01}. Our dataset comprises observations from 20 participants as one participant's data was excluded due to technical issues during data collection. 

\begin{table}[ht]
\centering
\caption{Summary of linear mixed models used for data analysis}
\label{LMM_factors}
\begin{tabular}{lp{5cm}}
\hline
\textbf{Dependent Variable} & \textbf{Independent Variables} \\ 
\hline
Resolution time & Interface (camera, VR, MR), error scenario variation (A, B, C), the number of objects left and the contextualisation time. \\
SUS Score & Interface (camera, VR, MR), error scenario variation (A, B, C), success of the attempt, and resolution time. \\
NASA TLX Score & Interface (camera, VR, MR), error scenario variation (A, B, C), success of the attempt, and resolution time. \\
\hline
\end{tabular}
\end{table}

A summary of the different LMM models used during our analysis is provided in Table~\ref{LMM_factors}. For each model, independent variables were included as fixed factors, and a random factor, where the participant ID serves as a random effect, was added to account for each participant being a random sample representative of the entire population. In our analysis, we tested for interactions in all our models, specifically investigating potential two-way interactions between the interface and error scenario variation and a four-way interaction among all independent variables. Using a significance threshold of $p \leq 0.05$, no significant interactions were detected across models. Consequently, we proceeded with the simpler models, which do not include any interaction terms.

For each final model, normality and constant variance assumptions were validated through the Shapiro-Wilk test and a visual inspection of Q-Q and scale location plots. Dependent variables were transformed when required to meet the aforementioned assumptions. We conducted post-hoc pairwise comparisons using the \textcolor{blue}{Tukey} test for the interface and scenario factors in the instances where they were found to be statistically significant. The Benjamini-Hochberg (BH) correction procedure was applied to control the false discovery rate for multiple comparisons as this adjustment helps to reduce the likelihood of Type I errors (i.e. finding a statistical difference when there is none) while maintaining statistical power. All analyses in this study considered a significance level $\alpha \leq 0.05$. 

\textbf{Resolution Time Analysis.} %
In this model, the number of objects left and contextualisation time were also included as independent factors. We observed that leaving objects behind led to a shorter resolution time as participants clicked finish without collecting all objects. Contextualisation time was included to account for variations in how long participants took before starting to pick a cube.

The square root of the resolution time was used as the dependent variable for this model. This square root transformation was done to ensure the constant variance assumption for the residual of the model was met~\cite{https://doi.org/10.2307/2983678}. This model has a cardinal variance\footnote{Cardinal variance is the proportion of the variance for the dependent variable explained by all fixed effects plus the participant random effect.} of \( R^2= 0.779\).

\begin{table}[ht]
\centering
\caption{Results of the Linear Mixed Model Analysis on Square Rooted Resolution Time, Controlling for Random Effects. $\beta$ coefficients indicate effect sizes.}
\label{LMM_resolve}
\begin{tabular}{p{2cm}p{1cm}rrrr}
\hline
\textbf{Predictor} & \textbf{Estimate $(\beta)$} & \textbf{Std. Error} & \textbf{t value} & \textbf{p value} \\ 
\hline
(Intercept)       & 15.564 & 0.314  & 49.508  & 2e-16 & *** \\ 
interfaceMR       & -0.610 & 0.235  & -2.601  & 0.014   & *   \\ 
interfaceVR       & -0.030 & 0.229  & -0.130  & 0.898   &     \\ 
scenarioB         &  0.059 & 0.224  &  0.263  & 0.794   &     \\ 
scenarioC         &  0.863 & 0.243  &  3.549  & 0.001   & **  \\ 
num. objects left    & -1.988 & 0.270  & -7.375  & 1e-09  & *** \\ 
contextualisation time      &  0.006 & 0.003  &  2.087  & 0.043   & *   \\ 
\hline
\end{tabular}
\begin{flushleft}
\textit{Signif. codes:  0 ‘***’ 0.001 ‘**’ 0.01 ‘*’ 0.05 ‘.’ 0.1 ‘ ’ 1}
\end{flushleft}
\end{table}

As shown in Table \ref{LMM_resolve}, the interface was found to have a statistically significant effect on the resolution time. Specifically, the MR interface obtained an estimated $\beta$ coefficient of -0.61 (-19.27 seconds), with a 95\% confidence interval ranging from -1.05 (-33.15 seconds) to -0.17 (-5.37 seconds). This indicates that the MR interface resulted in better performance as participants took less time to complete the task when using this interface. Post-hoc pairwise comparisons for this factor show a statistically significant difference in resolution time between MR and the other interfaces (VR and camera). On average, the resolution time was 19.27 seconds longer for the camera interface and 18.34 seconds longer for the VR interface compared to the MR interface resolution time. However, there was no significant difference in resolution time between the VR and camera interfaces. {\textcolor{blue}{These results are based on model-adjusted means, accounting for variability across conditions. The raw average time for each interface was 4 min. 16 sec. for the camera streams, 4 min. 15 sec. for VR, and 3 min. 57 sec. for MR.}}

The scenario was also found to have a statistically significant effect on the resolution time, with an estimated $\beta$ coefficient for scenario C of 0.86 (27.24 seconds), and a 95\% confidence interval ranging from 0.41 (12.94 seconds) to 1.32 (41.67 seconds)\footnote{We note that all our results show wide confidence intervals alongside small p-values. This can be attributed to the small sample size of 20 participants, however, even with this high variability in the data, significant effects can still be observed when effect sizes are relatively large \cite{https://doi.org/10.1111/j.1469-185X.2007.00027.x}.}. This result suggests that scenario C had higher complexity as participants took longer to complete it. The post-hoc comparisons for the scenario factor indicate a statistically significant difference in resolution time between scenarios A and C and scenarios B and C. On average, scenario C took 27.25 seconds longer than scenario A and 25.38 seconds longer than scenario B. No statistically significant difference was found between scenarios A and B. {\textcolor{blue}{These results are based on model-adjusted means, accounting for variability across conditions. The raw average time for each scenario was 4 min. 1 sec. for scenario A, 4 min. for scenario B, and 4 min. 32 sec. for scenario C.}}

As initially assumed, the number of objects left ($\beta$ = -1.99 (-62.82 seconds), with a 95\% confidence interval ranging from -2.49 (-78.61 seconds) to -1.48 (-46.72 seconds).) and contextualization time ($\beta$ = 0.01 (0.32 seconds), with a 95\% confidence interval ranging from 0.00 (0 seconds) to -0.01 (-0.32 seconds) were also found to have a statistically significant effect on the resolution time. This result confirms that attempts for which more objects were left behind resulted in lower resolution times, and higher contextualization times led on average to slightly longer resolution times.

\begin{figure}
    \centering
    \includegraphics[width=1\linewidth]{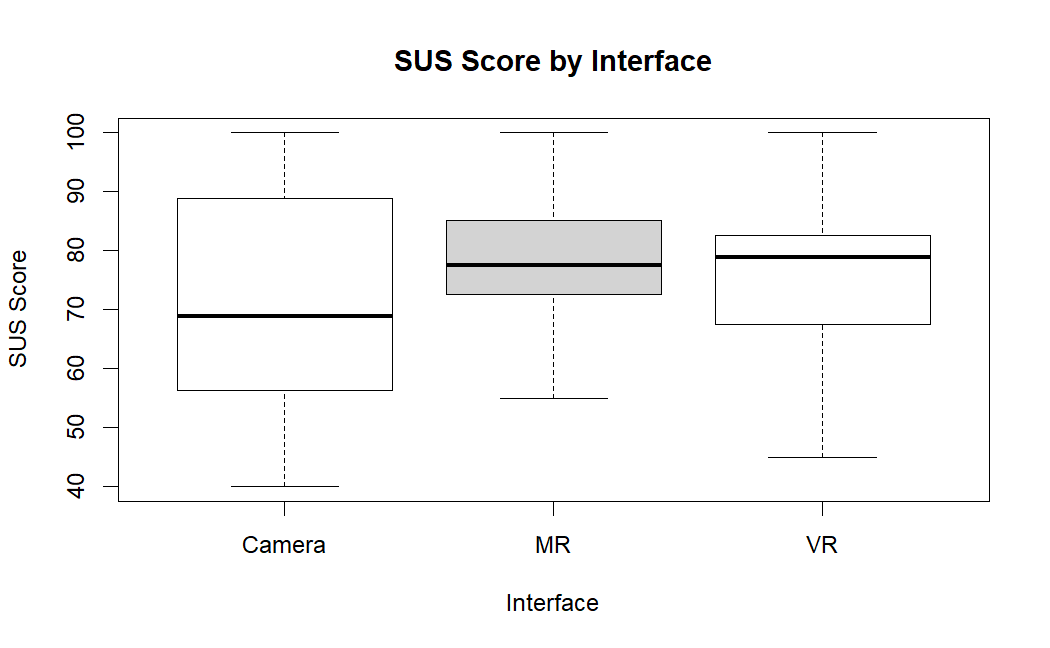}
    \caption{Distributions of usability scores for each interface}
    \label{fig:plot-sus}
\end{figure}

\textbf{SUS Score Analysis.} In this model, the success of the attempt and resolution time were included as independent factors. We observed that participants' perception of completing the task successfully and how long they took to do so impacted the SUS score calculated from their responses in the post-interface survey. This model has a cardinal variance of \( R^2 = 0.545 \). Table \ref{LMM_SUS} shows the model outcomes and Fig. \ref{fig:plot-sus} shows the usability scores for all interfaces.

\begin{table}[ht]
\centering
\caption{Results of the Linear Mixed Model Analysis for SUS Score, Controlling for Random Effects. $\beta$ coefficients indicate effect sizes.}
\label{LMM_SUS}
\begin{tabular}{p{1.3cm}p{1.2cm}rrrr}
\hline
\textbf{Predictor} & \textbf{Estimate $(\beta)$} & \textbf{Std. Error} & \textbf{t value} & \textbf{p value} \\ 
\hline
(Intercept)     & 60.112  & 12.474 & 4.819  & 0.00001  & ***  \\ 
interfaceMR     & 10.929  &  3.881 & 2.816  & 0.0079     & **   \\ 
interfaceVR     &  1.764  &  3.638 & 0.485  & 0.6307     &      \\ 
scenarioB       &  0.112  &  3.643 & 0.031  & 0.9756     &      \\ 
scenarioC       & -9.831  &  4.351 & -2.260 & 0.0293     & *    \\ 
success.fail    & -15.756 &  7.503 & -2.100 & 0.0405     & *    \\ 
resolution time    &  0.108  &  0.062 & 1.756  & 0.0848     & .    \\ 
\hline
\end{tabular}
\begin{flushleft}
\textit{Signif. codes:  0 ‘***’ 0.001 ‘**’ 0.01 ‘*’ 0.05 ‘.’ 0.1 ‘ ’ 1}
\end{flushleft}
\end{table}

Our results show that the interface had a statistically significant effect on participants' usability ratings. We observed an estimated $\beta$ coefficient for the MR interface of 10.93 with a 95\% confidence interval ranging from 3.66 to 18.17, indicating that participants perceived MR as easier to use compared to the other interfaces. The post-hoc pairwise comparisons for the interface factor show a statistically significant difference between the usability ratings of the camera and MR interfaces, with the camera interface SUS score being 10.93 points lower on average than the MR SUS score. Furthermore, the difference between the VR and MR interface usability scores is also statistically significant, with the VR interface SUS score being 9.16 lower on average than the MR interface SUS score. However, there was no significant difference in usability scores between the VR and camera interfaces.

The scenario also had a statistically significant effect on the usability scores, the estimated beta coefficient for scenario C is $\beta$ = -9.83 with a 95\% confidence interval ranging from -17.95 to -1.71. The post-hoc comparisons for the scenario factor indicate statistically significant differences between the SUS scores attributed to scenarios A and C, and scenarios B and C. Overall,  scenario C had a 9.83 lower SUS score on average than scenario A and 9.94 lower SUS score than scenario B. This suggests that after doing scenario C, participants' usability ratings were on average lower across all interfaces.

The success of the attempt also had a statistically significant effect on the SUS score, with an estimated $\beta$ coefficient of -15.76 and a 95\% confidence interval ranging from -30.55 to -1.42. This result confirms that the success of the attempt on average resulted in higher SUS scores. However, there was no significant difference in SUS scores due to the resolution time.

\begin{figure}
    \centering
    \includegraphics[width=0.9\linewidth]{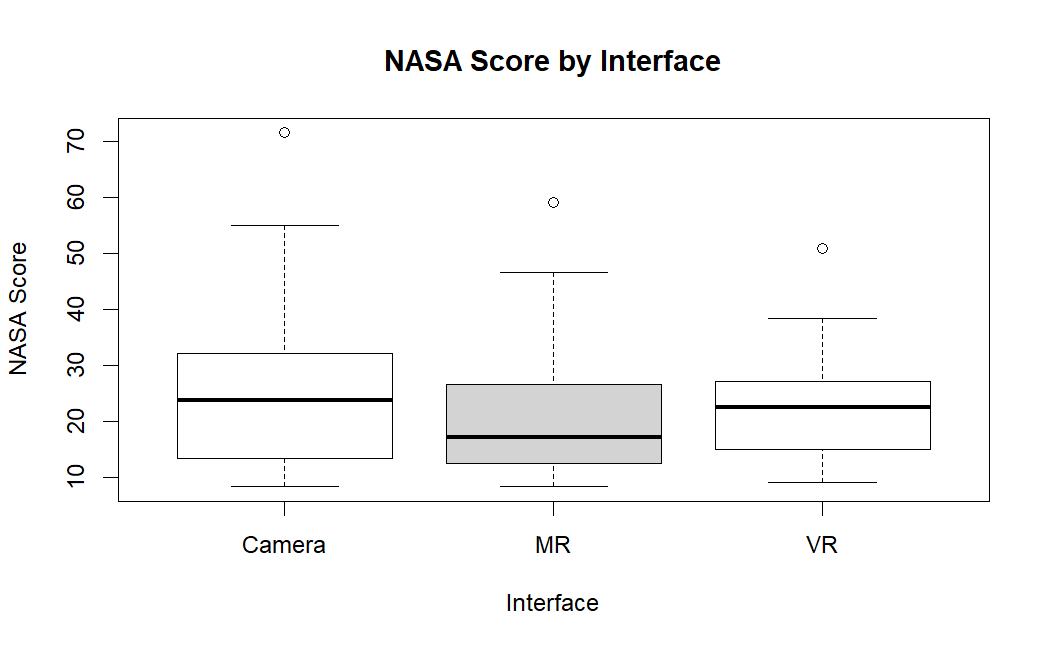}
    \caption{Distribution of Nasa TLX scores for each interface}
    \vspace{5pt}
    \label{fig:plot-nasa}
\end{figure}

\textbf{NASA TLX Score Analysis.} The square root of the NASA TLX score was used as the dependent variable for this model. This square root transformation was done to ensure the constant variance assumption for the residual of the model was met~\cite{https://doi.org/10.2307/2983678 }. Similar to the SUS score model, the success of the attempt and resolution time were also included as independent factors as we suspect that participants' perception of completing the task successfully and how long they took to do so might have impacted the NASA TLX score calculated from their responses in the post-interface survey. This model has a cardinal variance of \( R^2 = 0.816 \). Table \ref{LMM_NASA} shows the model outcomes, and Fig. \ref{fig:plot-nasa} shows Nasa TLX scores for all interfaces. {\textcolor{blue} {We used Residuals vs Leverage plots to assess the impact of Fig. 4's extreme values and found they were not influential in our linear regression fit, so we retained the participants' data points for analysis.}}

\begin{table}[ht] 
\centering 
\caption{Results of the Linear Mixed Model Analysis for NASA TLX Score, Controlling for Random Effects. $\beta$ coefficients indicate effect sizes} 
\label{LMM_NASA} 
\begin{tabular}{p{1.5cm}p{1cm}rrrr} 
\hline 
\textbf{Predictor} & \textbf{Estimate($\beta$)} & \textbf{Std. Error} & \textbf{t value} & \textbf{p value} \\ 
\hline 
(Intercept) & 5.101 & 0.818 & 6.234 & 9.4e-08 & *** \\ interfaceMR & -0.588 & 0.216 & -2.720 & 0.0102 & * \\ interfaceVR & -0.251 & 0.201 & -1.246 & 0.2214 & \\ scenarioB & 0.017 & 0.201 & 0.086 & 0.9321 & \\ 
scenarioC & 0.282 & 0.249 & 1.132 & 0.2649 & \\ success.fail & 0.160 & 0.469 & 0.341 & 0.7348 & \\ resolution time & -0.002 & 0.004 & -0.386 & 0.7015 & \\ \hline \end{tabular} \begin{flushleft} \textit{Signif. codes: 0 ‘***’ 0.001 ‘**’ 0.01 ‘*’ 0.05 ‘.’ 0.1 ‘ ’ 1} \end{flushleft}
\end{table}

As shown in Table \ref{LMM_NASA}, the interface had a statistically significant effect on the NASA TLX score, with an estimated $\beta$ coefficient for the MR interface of -0.588 (5.50 raw NASA score) and a 95\% confidence interval ranging from -0.99 (9.25 raw NASA score) to -0.19 (1.75 raw NASA score). This suggests that participants' perceived mental workload was lower when using this interface. Post-hoc pairwise comparisons for this factor show a statistically significant difference in NASA TLX scores between the MR and camera interfaces. On average, the NASA TLX score for the camera interface was 5.50 points higher compared to the MR interface. However, there were no significant differences in NASA TLX scores between VR and MR and camera and VR.

Finally, contrary to previous models, we found no significant differences in NASA TLX scores due to the error variation scenario or the success of an attempt.

\begin{figure}
    \centering
    \includegraphics[width=0.85\linewidth]{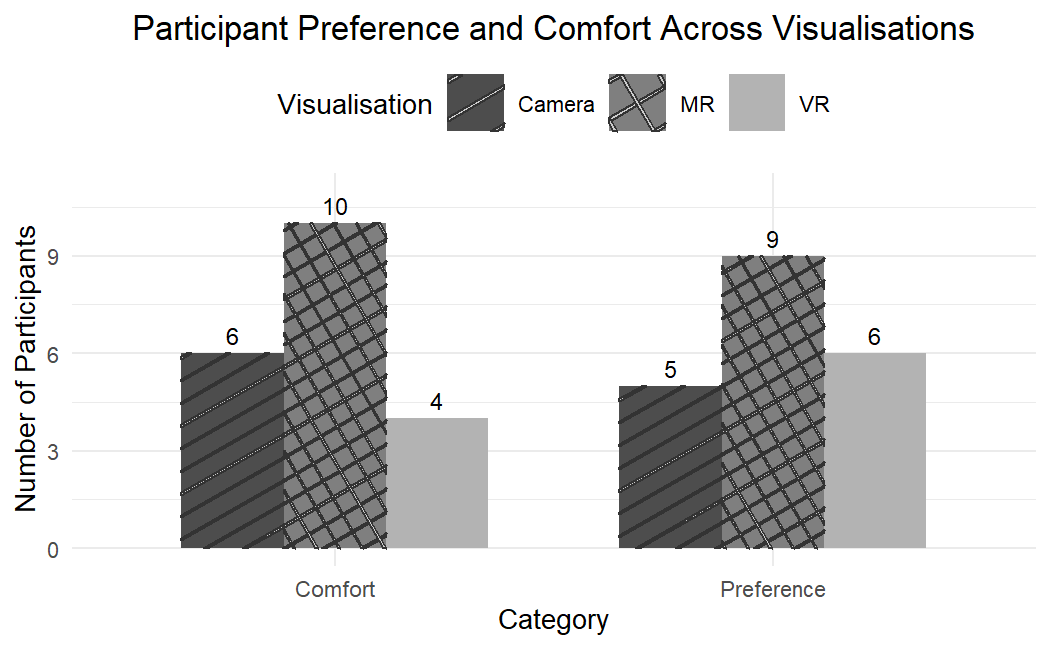}
    \caption{Post-study survey result regarding participants preferred each interface and which interface they found to be the most comfortable.}
    \label{fig:post-study-result}
\end{figure}

\textbf{Post-Study Survey Results.} Fig. \ref{fig:post-study-result} presents participants' responses regarding which interface they preferred and found to be the most comfortable for extended use. Among the interfaces, it can be seen that participants preferred MR, followed by VR, and then camera. In terms of comfort, more participants perceived MR as the most comfortable interface for extended use, followed by Camera Streams, with VR being rated as the least comfortable.

\section{Discussion}

MR and VR are offering new interaction mediums and interfaces for tasks and work related to remote error correction, especially in warehouse and distribution centre settings. In this study, we compared the performance and usability of MR, VR and a standard camera interface for a task emulating what remotely correcting an error in a warehouse may look like. 

The findings from both the objective and subjective metrics align with each other. The statistical analysis of the resolution time, SUS score and NASA TLX score, as well as participant feedback from the post-study questionnaire, all indicate that MR performed better, was perceived as the most usable and comfortable interface, and placed a lower cognitive load on participants during usage. 
This is further supported by participants' comments preferring MR over VR, with one participant specifically mentioning disorientation while using VR due to the lack of a physical reference point. Some participants also commented that certain interfaces made the task easier, while the camera interface, in particular, was perceived as more difficult to navigate.

This outcome was expected, as MR provides users with a virtual representation of the robot's environment while blending it with their physical surroundings. This combination enhances users' understanding of the robot's environment and increases their confidence when moving and investigating the scene, without the concern of bumping into objects in their physical space. Furthermore, the MR interface allowed participants to still see their own bodies and arms and research has shown that these cues are important for correct spatial understanding \cite{coello2005spatial}. 

Our statistical analysis supports hypotheses \textbf{H1}, participants were on average faster when using MR, and \textbf{H2}, participants attributed higher usability scores and lower NASA TLX scores to MR. These findings are consistent with prior research showing that interfaces incorporating, such as MR, both the real world and virtual visualisations have faster work completion than purely virtual interfaces (VR) \cite{7833028}. Hypothesis $H3$ is also supported by our data as the majority of the participants expressed a stronger preference for MR. Other works comparing MR and VR interfaces for tasks have had users prefer the VR interface \cite{app132111693}. However, the mentioned study used differing hardware for their VR and real-world and digital visualisation combination interface and attributes the users' preference to the handheld controllers more so than to the actual visualisation interface itself. 

Despite the data showing that MR is superior to VR and camera interfaces across all metrics, some participants still found VR more comfortable to use as they noted that the surroundings in MR appeared blurry. However, this could improve in the future as XR headsets become more advanced, providing clearer passthrough visuals with higher refresh rates and lower latency. Additionally, some participants commented on the inconsistency between the virtual table used in MR, which was circular, and the actual rectangular table of the robot, suggesting the addition of an outline of the robot's working space to better align with the real environment.

Lastly, although variations of the error scenario were tested to ensure similar complexity, our data suggests that scenario C may have been more complex, as it required more time to complete. This could be due to the need for greater exploration, particularly moving to areas behind the robot more frequently than in other scenarios. Participants also noted that the robot’s movement was slow and jittery, which likely caused delays when maneuvering. They suggested improvements such as more intuitive exploration control, smoother operation, covering more distance per control input, or automating some of the robot’s detection tasks. Additionally, participants recommended that detected objects should persist (remain visible even when not within the end-effector's view) in both VR and MR and that parts of the table already swept or visited should be marked or coloured to facilitate faster task resolution.

\section{Conclusion}
We compared three interfaces—MR, VR, and Camera Streams—for a remote error-resolution task. The task simulated a packaging error, where items were scattered across the workspace, requiring sorting - a common issue in robotic pick-and-place operations \cite{10160846}. A proof-of-concept application was developed for each interface, enabling users to control the robot's joints for exploration and use a click-and-go mechanism to pick and place cubes with the robotic arm. In a user study involving 21 participants, users were tasked with picking up cubes in the robot's environment and placing them in correctly labelled bins using each interface. We measured task completion time and the number of cubes left unplaced and gathered user feedback through the SUS and NASA TLX questionnaires.

Results showed that tasks were completed faster using MR, which was also perceived as more intuitive, reflected in its significantly higher usability scores compared to VR and Camera Streams interface. MR also had lower cognitive workload ratings than the camera interface, based on NASA TLX scores, and was preferred over both VR and Camera Stream in terms of usability and comfort. These findings highlight MR's potential as a valuable tool in industrial settings where quick, accurate human-robot collaboration is essential, particularly in error-prone environments.

Despite the promising results, there are limitations to our study design. First, the error scenario we used was simplified, focusing on one type of error, so the findings may not generalize to more complex real-world scenarios. We note however that {\textcolor{blue}{the error resolution process we exemplified in this scenario—such as examining the robot’s environment and assessing the object’s condition—remains applicable to other error contexts beyond our specific task.}} Second, the participant pool was limited to 21 university students, of which 19 identified as male, which may not provide a sufficiently diverse sample to draw strong conclusions. Finally, although efforts were made to ensure that the error scenarios were of similar difficulty, the results showed that one scenario was more complex and required more time to resolve. Combined with the slow and jittery robot movement, these factors may have impacted certain metrics in our user study.

Future work should explore more complex error scenarios to better reflect real-world tasks. Improving control mechanisms—such as developing more intuitive joint control and smoother robot movement—could also enhance user experience, as suggested by participant feedback. Additionally, recruiting a larger and more diverse participant pool, including actual remote robot operators, would provide more relevant insights into the efficacy of MR and other interfaces in industrial applications.

\bibliographystyle{ieeetr}
\balance
\bibliography{refs}

\begin{thebibliography}{10}

\bibitem{shaw2022online}
N.~Shaw, B.~Eschenbrenner, and D.~Baier, ``Online shopping continuance after covid-19: A comparison of canada, germany and the united states,'' {\em Journal of Retailing and Consumer Services}, vol.~69, 2022.

\bibitem{amazonWebsite}
Amazon, ``10 years of amazon robotics: how robots help sort packages, move product, and improve safety.,'' 2022.

\bibitem{amazon-sparrow}
A.~Staff, ``{A}mazon introduces {S}parrow—a state-of-the-art robot that handles millions of diverse products --- aboutamazon.com.''
\newblock [Accessed 26-09-2024].

\bibitem{ocado}
J.~Prisco, ``{W}hy online supermarket {O}cado wants to take the human touch out of groceries {C}{N}{N}- edition.cnn.com.''
\newblock [Accessed 26-09-2024].

\bibitem{walmart}
M.~Gray, ``{A} {F}ork in the {R}oad: {W}almart {B}ets on {A}ssociates, {A}utomation,'' 2024.
\newblock [Accessed 30-09-2024].

\bibitem{8895161}
J.~Ren {\em et~al.}, ``A deep learning method for multiple faults detection and classification of unmanned ground vehicles,'' in {\em Computer Science and Information Technologies (CSIT)}, 2019.

\bibitem{honig2018understanding}
S.~Honig and T.~Oron-Gilad, ``Understanding and resolving failures in human-robot interaction: Literature review and model development,'' {\em Frontiers in psychology}, vol.~9, p.~861, 2018.

\bibitem{8832130}
G.~Avalle {\em et~al.}, ``An augmented reality system to support fault visualization in industrial robotic tasks,'' {\em IEEE Access}, 2019.

\bibitem{wozniak2023happily}
M.~K. Wozniak {\em et~al.}, ``Happily error after: Framework development and user study for correcting robot perception errors in virtual reality,'' {\em arXiv preprint arXiv:2306.14589}, 2023.

\bibitem{dragan2013policy}
A.~Dragan and S.~S. Srinivasa, ``A policy-blending formalism for shared control,'' {\em The International Journal of Robotics Research}, 2013.

\bibitem{xr-review}
Y.-P. Su {\em et~al.}, ``Integrating virtual, mixed, and augmented reality into remote robotic applications: A brief review of extended reality-enhanced robotic systems for intuitive telemanipulation and telemanufacturing tasks in hazardous conditions,'' {\em Applied Sciences}, vol.~13, no.~22, 2023.

\bibitem{10.1145/3597623}
M.~Walker {\em et~al.}, ``Virtual, augmented, and mixed reality for human-robot interaction: A survey and virtual design element taxonomy,'' {\em J. Hum.-Robot Interact.}, 2023.

\bibitem{whitney2019comparing}
D.~Whitney {\em et~al.}, ``Comparing robot grasping teleoperation across desktop and virtual reality with ros reality,'' in {\em International Symposium of Robotics Research (ISRR)}, 2019.

\bibitem{gui-vs-vr}
J.~Chen {\em et~al.}, ``Comparing a graphical user interface, hand gestures and controller in virtual reality for robot teleoperation,'' in {\em ACM/IEEE International Conference on Human-Robot Interaction}, 2023.

\bibitem{shi2023research}
Y.~Shi {\em et~al.}, ``Research on mixed reality visual augmentation method for teleoperation interactive system,'' in {\em International conference on human-computer interaction}, Springer, 2023.

\bibitem{nakamura2020dual}
K.~Nakamura {\em et~al.}, ``Dual-arm robot teleoperation support with the virtual world,'' {\em Artificial Life and Robotics}, 2020.

\bibitem{waymouth2021demonstrating}
B.~Waymouth {\em et~al.}, ``Demonstrating cloth folding to robots: Design and evaluation of a 2d and a 3d user interface,'' in {\em IEEE International Conference on Robot \& Human Interactive Communication}, 2021.

\bibitem{sar-teleop}
C.~Cruz~Ulloa {\em et~al.}, ``A mixed-reality tele-operation method for high-level control of a legged-manipulator robot,'' {\em Sensors}, 2022.

\bibitem{cruz2024analysis}
C.~Cruz~Ulloa, D.~Dom{\'\i}nguez, J.~del Cerro, and A.~Barrientos, ``Analysis of mr--vr tele-operation methods for legged-manipulator robots,'' {\em Virtual Reality}, vol.~28, no.~3, p.~131, 2024.

\bibitem{adaptive-ar}
G.~Avalle {\em et~al.}, ``An augmented reality system to support fault visualization in industrial robotic tasks,'' {\em IEEE Access}, vol.~7, pp.~132343--132359, 2019.

\bibitem{robot-preception-error}
K.~N. Kaipa {\em et~al.}, ``Resolving automated perception system failures in bin-picking tasks using assistance from remote human operators,'' in {\em IEEE International Conference on Automation Science and Engineering (CASE)}, 2015.

\bibitem{10160846}
C.~Mitash {\em et~al.}, ``Armbench: An object-centric benchmark dataset for robotic manipulation,'' in {\em IEEE International Conference on Robotics and Automation (ICRA)}, 2023.

\bibitem{vr-camera}
B.~Omarali {\em et~al.}, ``Virtual reality based telerobotics framework with depth cameras,'' in {\em IEEE International Conference on Robot and Human Interactive Communication (RO-MAN)}, 2020.

\bibitem{9106630}
R.~Hetrick {\em et~al.}, ``Comparing virtual reality interfaces for the teleoperation of robots,'' in {\em Systems and Information Engineering Design Symposium (SIEDS)}, 2020.

\bibitem{coello2005spatial}
Y.~Coello, ``Spatial context and visual perception for action,'' {\em Psicologica: International Journal of Methodology and Experimental Psychology}, vol.~26, no.~1, pp.~39--59, 2005.

\bibitem{sus}
J.~Brooke, ``Sus: A quick and dirty usability scale,'' {\em Usability Evaluation in Industry}, 1995.

\bibitem{nasa}
S.~Hart, ``Development of nasa-tlx (task load index): Results of empirical and theoretical research,'' {\em Human Mental Workload/Elsevier}, 1988.

\bibitem{RProject}
``The r project for statistical computing.'' \url{https://www.r-project.org/}, 2024.
\newblock Accessed: 2024-10-02.

\bibitem{JSSv067i01}
D.~Bates, M.~Mächler, B.~Bolker, and S.~Walker, ``Fitting linear mixed-effects models using lme4,'' {\em Journal of Statistical Software}, 2015.

\bibitem{https://doi.org/10.2307/2983678}
M.~S. Bartlett, ``The square root transformation in analysis of variance,'' {\em Supplement to the Journal of the Royal Statistical Society}, vol.~3, no.~1, pp.~68--78, 1936.

\bibitem{https://doi.org/10.1111/j.1469-185X.2007.00027.x}
S.~Nakagawa {\em et~al.}, ``Effect size, confidence interval and statistical significance: a practical guide for biologists,'' {\em Biological Reviews}, 2007.

\bibitem{7833028}
M.~Krichenbauer {\em et~al.}, ``Augmented reality versus virtual reality for 3d object manipulation,'' {\em IEEE Transactions on Visualization and Computer Graphics}, vol.~24, no.~2, 2018.

\bibitem{app132111693}
N.~Kyaw {\em et~al.}, ``Comparing usability of augmented reality and virtual reality for creating virtual bounding boxes of real objects,'' {\em Applied Sciences}, vol.~13, no.~21, 2023.

\end{thebibliography}

\end{document}